\documentclass{article}
\usepackage{spconf,amsmath,epsfig}

\usepackage{bbm}
\usepackage{subfigure}
\usepackage{algorithm}
\usepackage{algorithmic}
\usepackage{multicol}
\usepackage{multirow}
\usepackage[american]{babel}
\usepackage{microtype}

\begin{document}
\title{AID++: An updated version of AID on scene classification}
%
\name{Pu Jin${}^1$, Gui-Song Xia${}^1$, Fan Hu${}^{1,2}$, Qikai Lu${}^{1,2}$, Liangpei Zhang${}^1$\thanks{This work was partially supported by NSFC projects under the contracts No.61771350 and No.41501462}}
\address{${}^1$ Key State Laboratory LIESMARS, Wuhan University, Wuhan 430072, China\\
${}^2$ Electronic Information School, Wuhan University, Wuhan 430072, China}

%

\maketitle
\begin{abstract}
Aerial image scene classification is a fundamental problem for understanding high-resolution remote sensing images and has become an active research task in the field of remote sensing due to its important role in a wide range of applications. However, the limitations of existing datasets for scene classification, such as the small scale and low-diversity, severely hamper the potential usage of the new generation deep convolutional neural networks (CNNs). Although huge efforts have been made in building large-scale datasets very recently, e.g., the Aerial Image Dataset (AID) which contains 10,000 image samples, they are still far from sufficient to fully train a high-capacity deep CNN model. To this end, we present a larger-scale dataset in this paper, named as AID++, for aerial scene classification based on the AID dataset. The proposed AID++ consists of more than 400,000 image samples that are semi-automatically annotated by using the existing the geo-graphical data. We evaluate several prevalent CNN models on the proposed dataset, and the results show that our dataset can be used as a promising benchmark for scene classification.
 \end{abstract}
\section{Introduction}
\label{sec:intro}
Scene classification is an important task in the interpretation of remote sensing image. In recent years, with the development of deep convolutional neural networks, deep CNNs have extraordinary performance in image classification tasks\cite{zhou2017places}. Transfer learning is regarded as a feasible method in the remote sensing image classification. Due to the great generalization of the CNN models pretrained on the large-scale data set ImageNet, the pretrained CNN models can achieve the great classification accuracy by trained on a few number of remote sensing images\cite{hu2015transferring}. However, the CNN models trained in this way are only applicable to the specific remote sensing data set cannot have a good performance on the other data set. The pretrained CNN models fine-tuned on a few number of remote sensing images are not completely suitable for remote sensing images and don't have good generalization ability. The lack of large-scale remote sensing image data set should be responsible for the bad generalization ability of the CNN models. The current remote sensing image scene classification data sets, such as AID\cite{xia2017aid}, UCM\cite{yang2010bag} cannot satisfy the demand of CNNs for the amount of data. So, there are few available scene classification datasets for training, testing, and comparing different scene classification algorithms. The lack of such a data set largely limits the development and application of scene classification algorithms. In order to alleviate these problems, we have created a data set containing 400,000 images and 46 categories. And we present the test results for state-of-the art convolution neural networks and evaluate the performance of different CNNs

In comparing the existing datasets\cite{xia2017aid}\cite{yang2010bag}, we found that the scene categories of existing datasets are very confusing, and the relationship between categories is not well organized. This confusion is very unfavorable to the semantic understanding of the image. So in response to this problem, we have created a hierarchical network of scene categories that organize categories clearly and effectively. Category networks have an important role in image annotation and image classification.The existing methods for establishing datasets mainly use manual annotation methods. However, when we want to create a dataset with a large amount of data, there is practically no feasibility only using manual annotation. We note that there are many kinds of data that describe surface features, such as maps, public geodata, and so on. These annotated information effectively describe the surface things. So, when we set up a data set, we can use these information to guide the annotation process of the image, which greatly reduces manpower consumption, improves annotation efficiency and ensures annotation accuracy.

This paper is organized as follows: We first describe the process of building a data set in Section 2, Section 3 compares the performance of the various convolutional neural networks on the data set, and finally draw the conclusion in Section 4.
\section{AID++ dataset}
\label{sec:index}

In the process of setting up a large-scale data set, the most time-consuming steps of the traditional man-made annotation method are to obtain the coordinates of scene categories and distinguish the scene types. Nowadays, there is a large amount of existing geo-information annotation in the remote sensing field. In the course of building the dataset, we can use map annotation information to find coordinates of a specific semantic tag, and then use the coordinates to get the corresponding images. The entire operation completes the data acquisition and labeling, which can greatly reduce annotation time and manpower consumption.
\begin{figure}[t]
	\centering
	\includegraphics[width = 1\linewidth]{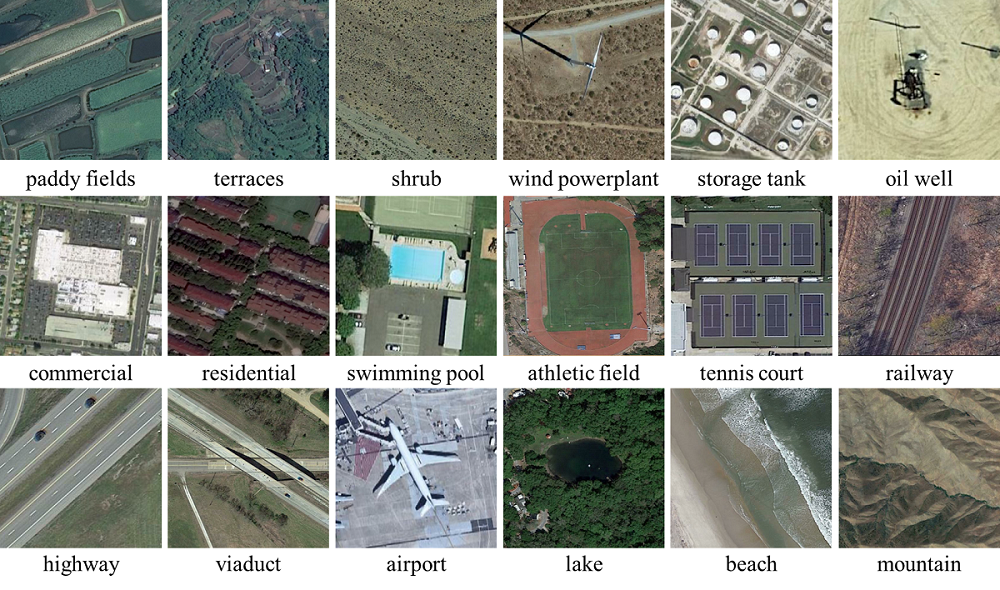}
	\caption{Samples of AID++: One sample of each semantic scene category is shown.}
	\label{fig:junc_detail}
\end{figure}

The construction of the AID++ dataset is composed of four steps, from creating a category network, using existing geodatabases to obtain coordinates of category, querying and downloading images using the coordinates, eliminating annotation errors manually, to scaling up the dataset and further improving the separation of similar classes.

\subsection{Category Network}
Because of the complexity and variety of remote sensing scene types, there are many kinds of relationships between categories. In order to effectively organize these scene categories, we have manually built an overcomplete multi-layered hierarchy for all scene categories. Through the semantic inclusion between scene categories, all remote sensing scene categories are arranged in a three-level tree: with 46 leaf nodes connected to 26 parent nodes at the second level that are in turn connected to 8 nodes at the first level (shown in Fig.1). By referring to various land-use and land-cover classification standards and selecting whether to meet the requirements by using aerial images, the selected categories are hierarchically organized by inclusion to form a final scene category network.

\begin{figure}[t]
	\centering
	\includegraphics[width = 1\linewidth]{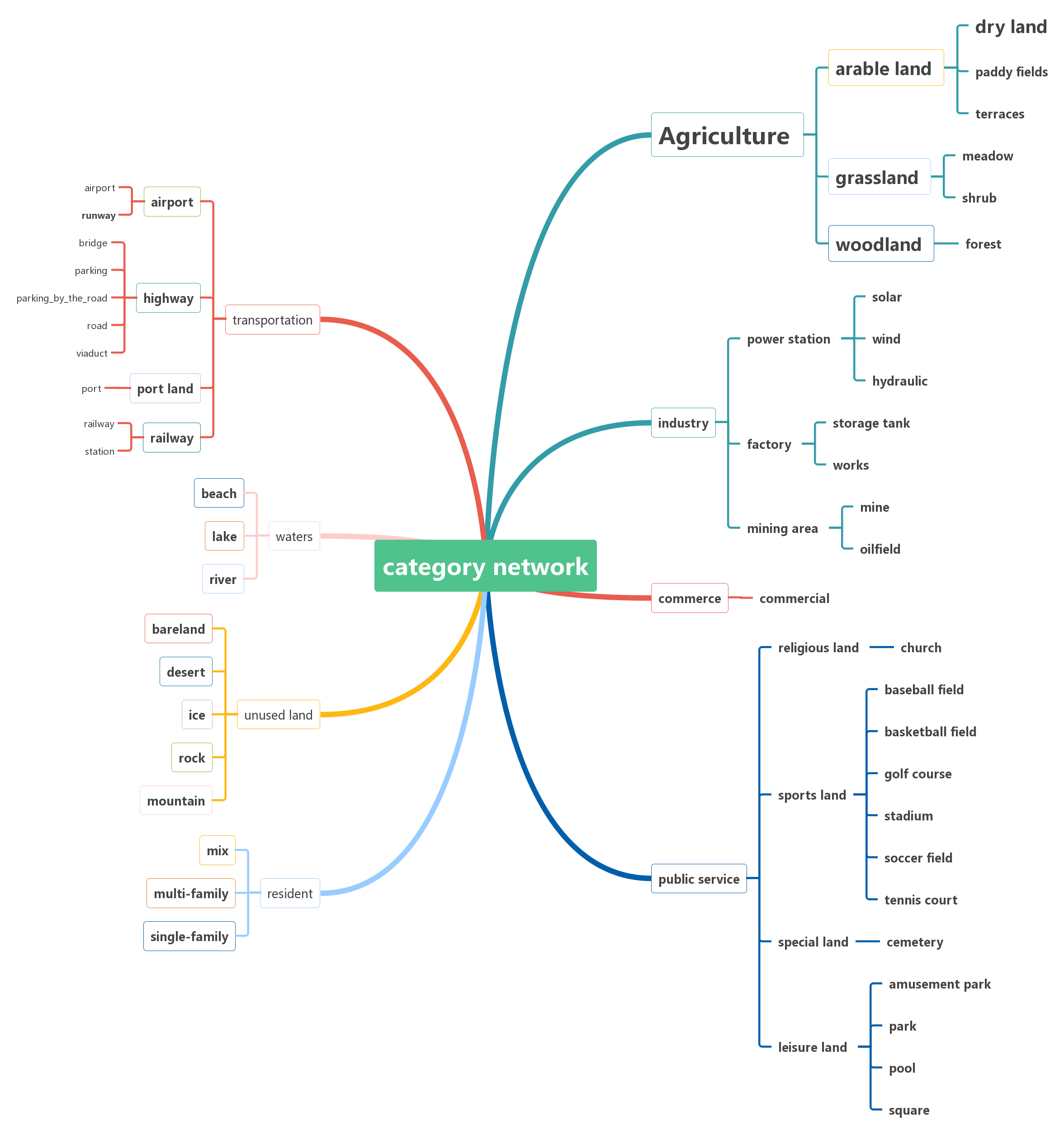}
	\caption{Category Network.}
	\label{fig:junc_detail}
\end{figure}

\subsection{Obtain Coordinates}
\label{sec:gbi}
After the construction of category network, the coordinates attached to the corresponding semantic tag should be collected by using existing geo-information annotation. We use three methods, Google Map API, OpenStreetMap, existing geodatabases, to obtain a large number of coordinates attached to a specific semantic tag. These coordinates will be used to obtain the corresponding image, thus completing the image acquisition and annotation. These methods take advantage of the information already available to annotate images, greatly reducing manpower requirements, and accelerating dataset expansion.
\begin{figure}[t]
	\centering
	\includegraphics[width = 1\linewidth]{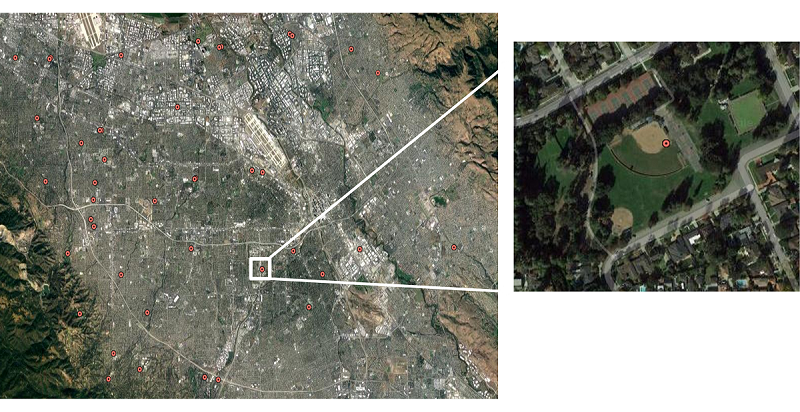}
	\caption{Left: The search result of basketball field. Right: The detail of the result.}
	\label{fig:junc_detail}
\end{figure}

\subsubsection{Google Map API}
The Google Map API is an application programming interface developed by Google Map. The API provides a map semantic tag search function. we use the API to develop a category tags search program, which can obtain the geographical coordinates matching tag information with a certain range. The result of the search program is displayed in Fig.2. The figure shows the result which is acquired by the program by searching the semantic tag “Baseball Field”

We search the label "baseball field", the program will search for all matching information points in the area, and will use the red dot to display the matching points. By changing the search area, the more matching points will be displayed and stored.
\subsubsection{OpenStreetMap}
OpenStreetMap is a collaborative project to create a free editable map of the world that allows registrants to freely edit maps under certain rules to add map information. The elements in the OpenStreetMap data model consist of node, way and relation(shown in Fig.4). The node is defined by its latitude, longitude and node id. The way is an ordered list of nodes which can be open or close. An open way is way describing a linear feature. Many roads, streams and railway lines are open ways. A closed way may be interpreted either as a closed polyline, or an area, or both.
\begin{figure}[t]
	\centering
	\includegraphics[width = 1\linewidth]{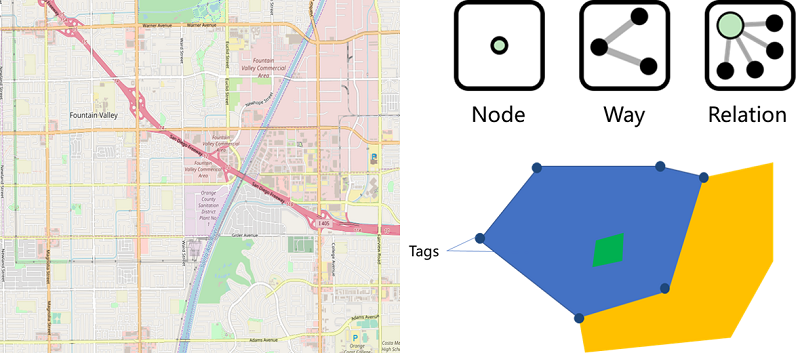}
	\caption{Left: The OpenStreetMap of a city. Right: The elements in the OpenStreetMap data model.}
	\label{fig:junc_detail}
\end{figure}

\begin{figure}[t]
	\centering
	\includegraphics[width = 1\linewidth]{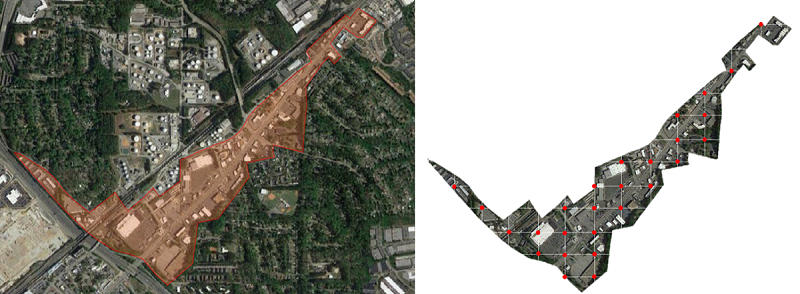}
	\caption{Left: A way outlines the commercial area. Right: The corrdinates are extract from the area.}
	\label{fig:junc_detail}
\end{figure}

Every node or way has tags which describe the surface feature. By searching the tag using the category, the nodes or ways which meet the search condition can be selected.  So, we will get the points or areas which have the semantic information matching with the category(shown in Fig.5).

By search “commercial” tag, we get a way which outlines the commercial area. So, we can get several images in the area which belongs to the category “commercial”.

\subsubsection{Existing Geodatabase}
Public existing geodatabases which are issued by state institutions have accurate and complete data. So those public databases which have excellent data are very important to the course of acquiring coordinates. Those databases provide the accurate coordinates and detailed information.

We obtained the National Bridge Inventory (NBI) Bridges dataset, which gives detailed information about the bridges in the United States, including coordinates, length, material, and more.
We get the coordinates of a large number of bridges by processing this dataset. Similarly, we also get other types of databases to get a lot of accurate coordinates.
We used to get the coordinates of a large number of bridges by processing this dataset. Similarly, we also get other types of data sets, so get a lot of high degree of confidence in the coordinate information.

\subsection{Acquiring images}
The images will be acquired by using the coordinates attached to the semantic tags. Taking the coordinates as the center, square boxes with a specific scale are formed to outline the areas which represent the scene categories. We will use the square boxes to delineate the scope and acquire the images in the scope. At last we will put the images which belongs to the same category together. But some images which have wrong annotation should be eliminated manual.
\section{Experiments}
\label{sec:experiment}	
\subsection{CNNs}
With the development of deep learning, convolutional neural networks perform well in remote sensing image scene classification. Several CNN models have excellent performance in scene classification tasks. VGG-16\cite{simonyan2014very} is a deep network developed based on AlexNet.The network increases the network layer to improve classification performance firstly. GoogleNet\cite{szegedy2015going} adopts the inception structure unit, which uses a parallel structure to parallelize the convolution kernels of different sizes. In order to solve the problem of gradient disappearance, ResNet\cite{he2016deep} sums the input and output of the layer. In this way, the input information is added directly to the output, so that the gradient is always present during gradient calculation of backpropagation. DenseNet\cite{huang2016densely} uses a densely connected block structure. The layers of the block are densely connected with each other. Compared with the ResNet, the connection of DenseNet don’t directly add the two sides.

\begin{table*}[]
\centering
\caption{Classification accuracy}
\label{my-label}
\begin{tabular}{|c|c|l|l|c|l|}
\hline
\multirow{2}{*}{Accuracy}                                & \multicolumn{5}{c|}{Training data set}                                                                                        \\ \cline{2-6} 
                                                         & NWPU                         & \multicolumn{1}{c|}{RSI-CB} & \multicolumn{1}{c|}{AID} & ImageNet & \multicolumn{1}{c|}{AID++} \\ \hline
\begin{tabular}[c]{@{}c@{}}Test on\\ WHU-19\end{tabular} & \multicolumn{1}{l|}{85.12\%} & 86.93\%                     & 90.45\%                  & 94.47\%  & 97.38\%                    \\ \hline
\end{tabular}
\end{table*}
\subsection{Results and analysis}

\begin{table}[]
\centering
\caption{Overall accuracy}
\label{my-label}
\begin{tabular}{|c|c|c|c|c|}
\hline
Networks         & VGG-16  & GoogleNet & ResNet  & DenseNet \\ \hline
OA & 76.58\% & 83.32\%   & 85.49\% & 86.61\% \\ \hline
\end{tabular}
\end{table}

We selected 46 categories in the category network for training and testing. Two kinds of proportion of images are used to train the CNNs. Table 1 shows the overall accuracy of the four networks. ResNet and GoogleNet have the almost same classification accuracy, while VGG-16 classification accuracy is the lowest. The DenseNet has the highest classification accuracy. The possible reason is that VGG-16 has fewer network layers than ResNet and does not extract more expressive representation of images\cite{hu2015transferring}. By comparing the results of different proportion of training images, we draw the conclusion that the more training images are, the higher classification accuracy is. 

By further studying the confusion matrix, we analyze the categories with high classification accuracy and poor classification accuracy. It can be found that the classification accuracy of land cover category can reach more than 90. However, the classification accuracy of land use category is lower. Categories under the same general category are often confusing.

In order to compare the generalization capabilities of different data sets training CNN models, we have designed multiple sets of transferring experiments. Several ResNet-50 models are trained on AID++, ImageNet\cite{deng2009imagenet}, NWPU\cite{cheng2017remote}, RSI-CB\cite{li2017rsi}, AID\cite{xia2017aid} respectively. And the pretrained models will be regarded to as a feature extractor classify the images of WHU-19, which is a small-scale remote sensing data set. The result of classification accuracy is displayed in the Table.2.

The model pretrained on AID++ achieve the best classification accuracy. Due to the larger-scale data, the model pretrained on AID++ achieve better classification accuracy than other remote sensing data sets. Comparing with ImageNet, the model pretrained on AID++ have a better generalization ability and achieve better classification accuracy. Hence, the model pretrained on AID++ is more applicable to the remote sensing scene classification.

\section{Conclusion}
\label{sec:conclusion}

In this paper, we present a new large-scale aerial image dataset for scene classification, for the sake of alleviating the current situation that the limitations of existing datasets considerably hamper the development of new high-power methods. Our benchmark dataset contains more than 400,000 images distributing in 46 classes. 

We propose an efficient method to label a large amount of image samples using the existing geo-databases. Several representative CNNs are trained and tested on the proposed dataset. The experimental results will be the promising benchmark for scene classification. Several comparative experiments demonstrate that the model pretrained on AID++ have a better generalization ability in the remote sensing scene classification than other remote sensing data sets and ImageNet. AID++ can be used to train the deep convolution neural networks and is more suitable for tasks in remote sensing field.

\footnotesize
\bibliographystyle{IEEEbib}
\bibliography{bib/refs}

\begin{thebibliography}{10}

\bibitem{zhou2017places}
Bolei Zhou, Agata Lapedriza, Aditya Khosla, Aude Oliva, and Antonio Torralba,
\newblock ``Places: A 10 million image database for scene recognition,''
\newblock {\em IEEE transactions on pattern analysis and machine intelligence},
  2017.

\bibitem{hu2015transferring}
Fan Hu, Gui-Song Xia, Jingwen Hu, and Liangpei Zhang,
\newblock ``Transferring deep convolutional neural networks for the scene
  classification of high-resolution remote sensing imagery,''
\newblock {\em Remote Sensing}, vol. 7, no. 11, pp. 14680--14707, 2015.

\bibitem{xia2017aid}
Gui-Song Xia, Jingwen Hu, Fan Hu, Baoguang Shi, Xiang Bai, Yanfei Zhong,
  Liangpei Zhang, and Xiaoqiang Lu,
\newblock ``{AID}: A benchmark data set for performance evaluation of aerial
  scene classification,''
\newblock {\em IEEE Transactions on Geoscience and Remote Sensing}, 2017.

\bibitem{yang2010bag}
Yi~Yang and Shawn Newsam,
\newblock ``Bag-of-visual-words and spatial extensions for land-use
  classification,''
\newblock in {\em Proceedings of the 18th SIGSPATIAL international conference
  on advances in geographic information systems}. ACM, 2010, pp. 270--279.

\bibitem{simonyan2014very}
Karen Simonyan and Andrew Zisserman,
\newblock ``Very deep convolutional networks for large-scale image
  recognition,''
\newblock {\em arXiv preprint arXiv:1409.1556}, 2014.

\bibitem{szegedy2015going}
Christian Szegedy, Wei Liu, Yangqing Jia, Pierre Sermanet, Scott Reed, Dragomir
  Anguelov, Dumitru Erhan, Vincent Vanhoucke, and Andrew Rabinovich,
\newblock ``Going deeper with convolutions,''
\newblock in {\em Proceedings of the IEEE conference on computer vision and
  pattern recognition}, 2015, pp. 1--9.

\bibitem{he2016deep}
Kaiming He, Xiangyu Zhang, Shaoqing Ren, and Jian Sun,
\newblock ``Deep residual learning for image recognition,''
\newblock in {\em Proceedings of the IEEE conference on computer vision and
  pattern recognition}, 2016, pp. 770--778.

\bibitem{huang2016densely}
Gao Huang, Zhuang Liu, Kilian~Q Weinberger, and Laurens van~der Maaten,
\newblock ``Densely connected convolutional networks,''
\newblock {\em arXiv preprint arXiv:1608.06993}, 2016.

\bibitem{deng2009imagenet}
Jia Deng, Wei Dong, Richard Socher, Li-Jia Li, Kai Li, and Li~Fei-Fei,
\newblock ``Imagenet: A large-scale hierarchical image database,''
\newblock in {\em Computer Vision and Pattern Recognition, 2009. CVPR 2009.
  IEEE Conference on}. IEEE, 2009, pp. 248--255.

\bibitem{cheng2017remote}
Gong Cheng, Junwei Han, and Xiaoqiang Lu,
\newblock ``Remote sensing image scene classification: benchmark and state of
  the art,''
\newblock {\em Proceedings of the IEEE}, vol. 105, no. 10, pp. 1865--1883,
  2017.

\bibitem{li2017rsi}
Haifeng Li, Chao Tao, Zhixiang Wu, Jie Chen, Jianya Gong, and Min Deng,
\newblock ``Rsi-cb: A large scale remote sensing image classification benchmark
  via crowdsource data,''
\newblock {\em arXiv preprint arXiv:1705.10450}, 2017.

\end{thebibliography}

\end{document}